\documentclass{sigchi}

\usepackage[T1]{fontenc}   
\usepackage{txfonts}
\usepackage{mathptmx}
\usepackage[pdflang={en-US},pdftex,breaklinks=true,draft]{hyperref}
\usepackage{color}
\usepackage{booktabs}
\usepackage{textcomp}

\usepackage{microtype}        

\def\plaintitle{Automatic Player Identification in Dota 2}

\def\emptyauthor{}
\def\plainkeywords{Machine learning; online games; identification; digital forensics;}

\makeatletter
\def\url@leostyle{%
  \@ifundefined{selectfont}{
    \def\UrlFont{\sf}
  }{
    \def\UrlFont{\small\bf\ttfamily}
  }}
\makeatother
\def\pprw{8.5in}
\def\pprh{11in}

\definecolor{linkColor}{RGB}{6,125,233}
\hypersetup{%
  pdftitle={\plaintitle},
  pdfauthor={\emptyauthor},
  pdfkeywords={\plainkeywords},
  pdfdisplaydoctitle=true, 
  bookmarksnumbered,
  pdfstartview={FitH},
  colorlinks,
  citecolor=black,
  filecolor=black,
  linkcolor=black,
  urlcolor=linkColor,
  breaklinks=true,
  hypertexnames=false
}

\usepackage{sizhe}

\makeatletter
\def\@copyrightspace{\relax}
\makeatother
\begin{document}

\title{\plaintitle}

\numberofauthors{3}
\author{%
  \alignauthor{Sizhe Yuen\\
    \affaddr{Maritime Engineering}\\
    \affaddr{University of Southampton, UK}\\
    \email{sizhe1007@gmail.com}}\\
  \alignauthor{John D. Thomson\\
    \affaddr{School of Computer Science}\\
    \affaddr{University of St Andrews, UK}\\
    \email{j.thomson@st-andrews.ac.uk}}\\
  \alignauthor{Oliver Don\\
    \email{don@olli.me.uk}}\\
}

\maketitle
\begin{abstract}
  Dota 2 is a popular, multiplayer online video game. Like many online games, players are mostly anonymous, being tied only to  online accounts which can be readily obtained, sold and shared between multiple
  people.  This makes it difficult to track or ban players who exhibit unwanted behavior online.
  In this paper, we present a machine learning approach to
  identify players based a `digital fingerprint' of how they play the game, rather than by account.
  We use data on mouse movements, in-game statistics
  and game strategy extracted from match replays and show that for best results, all of these are necessary. We are able to obtain an accuracy of prediction of 95\%  for the problem of predicting if two different matches were played by the same player. 
\end{abstract}

\ccsdesc[500]{Information systems~Massively multiplayer online games}
\ccsdesc[500]{Applied computing~Computer games}
\ccsdesc[300]{Computing methodologies~Machine learning approaches}

\keywords{\plainkeywords}

\section{Introduction}

Dota 2 is a popular, multiplayer online video game. Like many online games, players are mostly anonymous, being tied only to  online accounts which can be readily obtained, sold and shared between multiple people. This paper presents an automatic way of identifying a player by establishing a `digital fingerprint' based on how a player plays the game rather than relying on account information. It shows that incorporating  game-centric information into the model is necessary for the best performance, when compared to existing context free digital forensic techniques which rely on mouse movements. 
  
Using a combination of game information and mouse movement analysis, we are able to identify if two different matches were played by the same player  with an an accuracy of 95\%.
  
We believe this is an interesting result in itself, but it also has real world utility. The anonymity of online gaming and the ease of obtaining new accounts often leads to unwanted behavior by a minority of players. This can manifest in a number of ways: abusive communication, cheating and matchmaking manipulation. Unchecked abusive communication often manifests itself in terms of sexist and homophobic abuse, which leads to an unpleasant and hostile game environment. Matchmaking manipulation occurs when a player uses a different account (usually purchased illegally) with a different skill rating than the player has themselves. This leads to unbalanced games. 
  
This presents a challenge for the developer of these games: simply banning accounts often does not work. As Dota 2 is a free to play multiplayer game, identifying players is especially challenging. A single person can create multiple accounts anonymously merely by providing different email addresses. As a result, unwanted displayed by some players towards one another \cite{toxic} and the rise of illegitimate secondary markets to sell accounts with higher ranks present a serious problem. Players who are banned from the game for bad behavior may continue to play using a fresh account. Additionally, without the ability to verify and link a person to an account, it is easy for players to cheat in amateur online tournaments by playing in the place of another player. A lot of manual effort and luck is required to determine if someone cheated this way. In a recent example, a player was caught cheating in the \$30,000 CSL Dota 2 tournament for college students. The cheater was only caught after officials were tipped off \cite{dota-cheating} due to obviously strange behavior. By using machine learning techniques to identify the players based on their behavior, cases like this would be easily identifiable and solvable.

Learning player identity by their behavior in order to take
action against them is also more effective compared to
taking action based on account names, IP addresses or hardware ID
bans. Players can just buy or create new accounts if their
original one is banned; IP addresses are rarely static
\cite{dynamic-ip} due to NAT gateways and banning a range of
address may lead to banning innocent players; Hardware ID bans
may be effective in the short term, but hardware can still
change hands between people. A system which
can recognise player behavior can be more effective in
the longer term, especially as we explore from a low-level
and fine-grained view of the data rather than high-level
behavioral elements.

\section{Background}

\subsection{Dota 2}

\begin{figure}
  \centering
  \includegraphics[width=0.7\columnwidth, keepaspectratio]{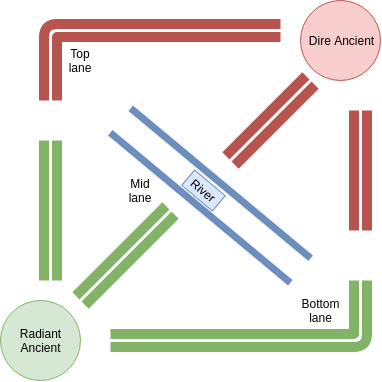}
  \caption{Representation of the map in Dota 2.}
\end{figure}

Defence of the Ancients (Dota) 2 \cite{dota} is a popular free-to-play multiplayer online
battle arena (MOBA) video game developed by Valve Corporation.
In Dota 2, the player controls a single hero as part
of a team of five players competing against an opponent team. Each hero has their own unique
abilities and characteristics, for example some heroes focus on dealing damage with spells and
abilities, while other heroes use passive abilities to augment their attacks. To win, the team
must destroy the opponent's base structure, known as the Ancient, located deep in the opponent
controlled side of the map. The map is split into three lanes, with towers that must be destroyed
in succession. Weaker, uncontrollable units called creeps periodically spawn from a team's
Ancient and travel down the three lanes. Heroes gain experience when nearby creeps and heroes on
the opposing team die. A hero then gains levels with experience, increasing the hero's powers and
abilities. 

Additionally, heroes gain gold throughout a match which can be used to purchase items. Items
provide powerful bonuses and abilities to a hero. Every hero has 6 slots in their inventory in
which they can place purchased items. Items can also be placed in the stash or backpack, but do
not provide any bonuses when placed there. Gold is gained from multiple sources, most notably by
getting the killing blow on creeps, killing opponent heroes, destroying opponent buildings and
passively over time.

The map in Dota 2 is asymmetrical and split in half for the two teams: Radiant and Dire.
Figure 1 shows a representation of what the map looks like. Playing
on different sides leads to slightly different strategies due to the asymmetry. The explanation
of other more complicated mechanics in the game, such as the neutral jungle creeps and runes are
omitted as they do not form any part of the features or data extracted in this work. 

\subsection{Related work}

\subsubsection{Starcraft 2 player identification}
The problem of player identification based on play style and behavior has been explored before
in the game of Starcraft 2. Liu et al. \cite{starcraft-identification} first looked at using
machine learning algorithms to identify a Starcraft 2 player from features extracted from match
replays and followed up with further research \cite{starcraft-actions} on predicting a player's
next actions based on their previous actions in the match.

Their work has a similar goal to ours to predict players based on behavioral data. However,
our work differs in multiple ways. Firstly, the gameplay for Starcraft 2 and Dota 2 are different
as the player controls a large number of units in the former and only one in the latter.
This means the features used for player prediction in Starcraft focused more heavily on
the strategy and build order of players rather than fine grained data such as mouse movement.
Secondly, Liu et al. focussed on the classification from a fixed pool of players. While we
also do the same, we go further and look at classifying pairs of matches belonging to the
same player. This allows our predictions to be much more general and not need to be re-trained
on every new pool of players in the dataset.

\subsubsection{Dota 2}
There has been a variety of previous work in data analysis on Dota 2. Much of the growing
research in Dota 2 comes from the wealth of data provided from the replay system, which ranges
from the positions of each player's mouse cursor to the health points of each creep entity.

One area of exploration with Dota 2 data is the classification of player roles. As Dota 2 is a
team game, each player on the team typically fulfils a certain role in a match, similar to
traditional sports. Gao et al. \cite{dota-gao} presented positive results in the
identification of \textit{both} the hero and role that a player played using a mix of performance
data and behavioral data that involved ability and item usage. Eggert et al. \cite{dota-eggert}
took a further step in feature generation for role classification by constructing complex
attributes using low-level data from a parsed replay of a match. This included features such as
player positional movement and damage done during teamfights.

Data analysis in Dota 2 does not only involve analysis of in-game data using replays. Hodge et
al. \cite{dota-mixed-rank-win} used a mix of pre-match draft data such as the history of a
player's performance on different heroes and in-game data of game states at different sliding
window intervals to predict the outcome of unseen matches. Taking
match outcome prediction further, Yang et al. \cite{dota-yang} were able to predict match
outcomes in real-time as the match progressed, to get the probabilities of each team winning at
every minute of a match. For real-time data, they showed data from later stages of a match were
much more informative, as early portions of a match were often similar to other matches. Their
work was compared with work from Conley and Perry \cite{dota-conley}, whose focus was in creating
an engine for hero pick recommendations based on win predictions, and with work from Kinkade and
Lim \cite{dota-kinkade}, who also investigated combinations of in-game statistics and draft picks
outcome prediction. Drachen et al. \cite{dota-skill} studied how team behavior varied as a
function of player skill, specifically on the movement and positioning of heroes by players as
spatial-temporal data. This contrasts with most other studies, which have focused on only
temporal data like the gold and XP per minute statistics.  

There is also existing research that uses qualitative data rather than quantitative data.
Nuangjumnong and Mitomo \cite{dota-leadership} conducted a survey on players which showed
correlation between the leadership style of players and the role they played in the team.
Summerville et al. \cite{dota-draft} studied the drafting phase of the game to find common trends
and predict the draft sequence. In particular, because the draft sequence was described as a list
of words, they chose to use a one-hot encoding with a categorical cross-entropy loss to encode
the hero names.

\subsubsection{Other video games}
Player behaviour in video games has often been modelled to gather meaningful data for
further analysis. Pfau et al. \cite{lineage} applied machine learning techniques
to construct deep player behaviour models in the video game Lineage II.
These models can be applied in online crime detection for cheating and for autonomous
testing of the game.
Bakkes et al. \cite{behaviour-model} provided approaches to player behavioural
modelling with the goals of improving game AI and improving game development.
Drachen et al. \cite{destiny} developed in-depth behavioural models for
players in the video game Destiny to identify distinct play-styles and profiles
players can be clustered into.

\subsubsection{Mouse dynamics}
In research unrelated to Dota 2, many studies have investigated the area of using mouse
dynamics to detect user identity. This is a behavioral biometric technology that analyses
different mouse movement attributes, such as the velocity or curvature of the mouse cursor. A
significant portion of Dota 2's gameplay revolves around the use of mouse clicks and movements,
combined with occasional key presses. Bhatnagar et al. \cite{mouse-vs-keyboard} compared the use
of mouse and keyboard dynamics as a biometric technique and stated that keystroke dynamics have
lower predictive accuracy, but mouse dynamics data tends to be inconsistent and error prone due
to the varied hardware input devices.

Mouse movement data extraction can be done explicity by using a predefined activity,
or implicity by monitoring typical activity without a specified task \cite{mouse-dynamics}. For
example, Gambao and Fred \cite{mouse-features} used a simple memory game to extract their data.
This method allows for well-defined actions. Feher et al. \cite{mouse-dynamics} went further by
categorising mouse movement features into distinct hierarchies, which allowed for precise
categorisation of various different types of mouse movement. Their results were compared with
a histogram based method from Awad and Traore \cite{mouse-histogram} of aggregating
different types of mouse action and obtained noteworthy improvement in the accuracy of their
models. 

\section{Methodology}
In this paper, we investigated the use of three different featuresets to determine who
the player is. Each featureset explores a different aspect of the game, to see which
aspects have a greater effect in predicting players.
The mouse movement features are fine-grained
biometric features that directly capture player behavior based on how they move their mouse.
The game statistics features are higher level results of player behavior.
For example, an aggressive player may tend towards higher kill/death statistics compared
to a more passive player. Finally, the itemization features capture both strategy and
gameplay behavior of players. Players typically have strategies (or builds) that
they use on the same hero which is reflected in the items they buy.
The position of the items in the inventory also indicate specific player behavior based
on their personal keyboard bindings for the game.

The features are used both individually and in combination with the other features on
three different machine learning models (logistic regression, random forest classifier
and multi-layer perceptron) to find the best model and feature combination for
predicting player behavior.

\subsection{Data collection and processing}
A combination of the OpenDota \cite{opendota} API and Valve's official API was used to download
replays. In particular, the OpenDota API allowed many aspects of the replays to be controlled,
such as the game mode and hero id, leading to less random variables. The following list of
parameters were used for our datasets:
\begin{itemize}
\item \textit{Game mode}: All pick
\item \textit{Team}: Radiant
\item \textit{Date}: Before 18 November 2018 (to avoid changes from patch 7.20 affecting
  some of the existing work)
\item \textit{Hero id}: Constant as the same hero was used for all the datasets
\end{itemize}

Match replays were parsed using \textit{clarity}, an open source parser developed by Martin
Schrodt \cite{clarity}. It allowed fine grained data to be extracted from each replay on each
game tick. After the data was extracted, more processing was done to encode the data into
useful features to create our dataset.

\subsection{Mouse movement features}
The mouse movement features extracted from the Dota 2 match replays followed the concept by
Feher et al \cite{mouse-dynamics} on forming a hierarchy of atomic actions such as left click
or mouse move and complex actions made up of multiple atomic actions, such as mouse drag.
As the data available from replays are less granular and do not contain specific data such as
mouse click down and mouse click up, the addition of keyboard presses were combined to form
the complex mouse actions in this work. Two classes of mouse actions are defined:
\textbf{atomic} and \textbf{complex}.

\subsubsection{Atomic mouse actions}
The four atomic mouse actions are as follows:
\begin{enumerate}
  \setlength\itemsep{-0.25em}
  \item Mouse movement sequence
  \item Attack command
  \item Move command
  \item Spell cast command
\end{enumerate}
A mouse movement sequence is defined as a sequence of positions. Rather than using a fixed
interval of time, a threshold $\tau$ is defined to end the sequence if no change in cursor
position has occurred within the threshold time. This more naturally records a sequence of
mouse movements to not break up a sequence in order to fit within a fixed time interval.
A larger threshold increases the average number of actions in the sequence.
A threshold of 500 milliseconds was used in Feher et al.'s \cite{mouse-dynamics} study.
This was reduced to 300 milliseconds for capturing Dota 2 mouse features as it was found a
500 millisecond threshold created fewer and longer mouse movement sequences than expected. 

Each movement sequence consisted of three vectors of length $n$, where $n$ is the number
of game ticks:
\begin{itemize}
\item $\boldsymbol{t} = \{t_i\}^{n}_{i=1}$ - The game tick
\item $\boldsymbol{x} = \{x_i\}^{n}_{i=1}$ - The x coordinate sampled on game tick $t_i$
\item $\boldsymbol{y} = \{y_i\}^{n}_{i=1}$ - The y coordinate sampled on game tick $t_i$
\end{itemize}
$n$ can vary between different movement sequences for longer or shorter sequences.
These vectors are further processed to give the list of mouse movement features shown
in table \ref{tbl:mm-features}, based on Gambao and Fred's \cite{mouse-features} features.

\begin{table}
\renewcommand*{\arraystretch}{2.5}
\centering
\begin{tabular}{ l r }
\textbf{Feature} & \textbf{Definition} \\ \hline
Angle of movement & $\theta_i = arctan(\dfrac{\delta y_1}{\delta x_1}) + \sum\limits_{j=1}^{i} \delta \theta_j$ \\ 
Curvature & $c = \dfrac{\delta\theta}{\delta s}$ \\ 
Rate of change of curvature & $\Delta c = \dfrac{\delta c}{\delta s}$ \\ 
Horizontal velocity & $V_x = \dfrac{\delta x}{\delta t}$ \\ 
Vertical velocity & $V_y = \dfrac{\delta y}{\delta t}$ \\ 
Velocity & $V = \sqrt{\delta V_{x}^{2} + \delta V_{y}^{2}}$ \\ 
Acceleration & $V' = \dfrac{\delta V}{\delta t}$ \\ 
Jerk & $V'' = \dfrac{\delta V'}{\delta t}$ \\ 
Angular velocity & $w = \dfrac{\delta \theta_t}{\delta t}$ \\ 
\end{tabular}
\caption{Basic mouse movement features.}
\label{tbl:mm-features}
\end{table}
The other three atomic actions consist simply of the game tick, x and y coordinate.
They represent in-game intent of the mouse movement. For example if a player right-clicks
on the terrain, it is a move action to move their character, but if a player right-clicks
an enemy entity, it is an attack action.

\subsubsection{Complex mouse actions}
The combination of a mouse movement sequence and one of the three command actions creates a
complex mouse action. There are three complex mouse actions, each corresponding to their
respective atomic command. Complex mouse actions represent a mouse movement sequence
leading up to a command. Each of the three complex actions are recorded in the same way, but
record different kinds of data. For example, the complex move action will be recorded throughout
a match, while the complex spell cast action will typically occur only during teamfights.

In addition to the features mentioned in table \ref{tbl:mm-features}, two more features are
included for complex mouse actions:
\begin{itemize}
  \item $t_n$ - The number of game ticks between the last mouse position of the movement sequence
  and mouse position of the command.
  \item $d$ - The distance travelled between the last mouse positions of the movement sequence
  and mouse position of the command
\end{itemize}

Finally, because the feature vectors for a mouse movement sequence are of varying length, some
statistical values must be taken. The minimum, maximum, mean and standard deviation are used
leading to a total of 38 features for complex mouse action.

\subsection{Game statistic features}
In-game statistics are a useful way to determine player performance in Dota 2. Many community
websites \cite{dotabuff, opendota} track these statistics to let players view the performances
of their match history. Many existing studies
\cite{dota-mixed-rank-win, dota-kinkade, dota-pu-yang, dota-yang} also use these features,
especially for predicting performance of teams and players.

The statistics extracted for this work are:
\begin{multicols}{2}
  \begin{itemize}
    \item Kills
    \item Assists
    \item Deaths
    \item Gold per minute
    \item XP per minute
    \item CS (creep score) per minute
    \item Denies
    \item Actions per minute
    \item Move commands on target per minute
    \item Move commands on position per minute
    \item Attack commands on target per minute
    \item Attack commands on position per minute
    \item Spell cast commands on target per minute
    \item Spell cast commands on position per minute
    \item Spell cast commands with no target per minute
    \item Hold position commands per minute
  \end{itemize}
\end{multicols}

Most of the statistics are taken as per minute average rather than total,
because they can vary greatly depending on
the length of the game. In general, the game statistics are strong indicators of performance
as winning teams usually have higher gold and XP numbers. They can also provide
behavioral information, for example aggressive players may often get more kills than
more passive players, they may get more deaths as a result as well. Further, there are a few
different methods for players to accomplish the same action, which can be revealed based on
whether they use commands on targets or on positions.

\subsection{Itemization features}

The items a player buys on their hero had been shown by Eggert et al. and Gao et al. as a
strong indicator of the hero selection and role of the player. By itself, itemization is likely
not enough to tell the difference between players, as multiple players on the same hero are
very likely to buy the same items. However, items can be placed into different inventory
positions, which becomes a more personal choice by players. Moreover, many items in Dota 2 have
active abilities. The inventory positions of the item therefore can reflect a player's
keyboard bindings. It is not possible to extract a player's personal settings from a replay,
so using item positions are a close approximate. Figure 2 shows how the items are placed in
different slots with associated key bindings.

\begin{figure}
  \centering
  \includegraphics[width=0.6\columnwidth]{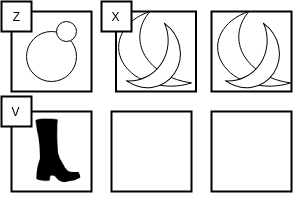}
  \caption{Representation of an inventory with items in Dota 2.
    The letters next to the items are bound to that inventory position
    and indicate the key to press to activate that item.}
  \label{fig:inventory}
\end{figure}

The inventories are sampled once at the beginning of a match and once at the end to prevent
variance. Items sampled at the beginning will not be affected by variables such as
performance as the game has just started. Items sampled at the end are similar, as most
items that players want will have already been purchased and are less affected by the flow
of the match.
The items and their positions are then encoded with multiple methods: feature hashing,
one-hot encoding, selective one-hot encoding and item difference.

Selective one-hot encoding only used a small subset of items deemed more useful for player
prediction. Two subsets were chosen, starting items and boots. There are only a small
number of items a player can buy at the start of a match, which restricts the number of items
from about 300 to 30 when looking at only starting items. This reduces the large number of
features that one-hot encoding generates ($\sim$ 300 items $\times$ 6 inventory locations) to
a reasonable number. Boots are a special case of items as almost all heroes will buy some
form of boots. There are 6 types of boots in the game, most of which have an active ability.
Because different boots have different abilities, but all give an important movement speed
increase the player characters, different players are likely to choose different types
of boots based on their personal play-style. Furthermore, the active ability makes the position
of the boots in the inventory an important attribute we can take advantage of.
This allowed for precise identification of both the type of boots and the inventory
positions they occupy as data.

\section{Experiments}
Two different experiments were done to explore the problem of predicting player identity
in Dota 2 based on the features discussed earlier. A dataset
comprising of 93 players with 356 matches was created. There was a mix of
number of matches per player, with some players having 10 matches in the
dataset while others only had 1 or 2 matches. The goal was to see if
we could successfully identify one player out of the pool of players in the dataset
given a single match.

The second experiment took a more generic approach. Rather than predict out of a static
pool of players, the goal was to identify if two matches belonged to the same player
or not. In this approach, there is no concept of individual players, only the binary
choice of whether the two given matches were played by the same person. A combination
of pairs of matches were created from the original dataset, which gave 3071
combinations where two matches belonged to the same player. The number of combinations
where two matches belonged to different players was much larger, but was randomly
filtered down to 3071 to ensure no bias towards the negative samples.

The features used in the second experiment
differed slightly from the first experiment because all matches have different
lengths, making it difficult to compare the features from two matches without further
modification. Further statistics are calculated for each mouse movement feature over
the entire match. For example, the horizontal velocity of a mouse action creates four
features: the mean, standard deviation, minimum and maximum horizontal velocity of
each \textit{individual} mouse action. There are many drawbacks to this approach,
most notably the loss of data from taking an average of averages, as the thousands
of mouse actions are reduced to only a small number of statistics. To alleviate this
issue, matches are split up into portions for each portion to have its own statistics.
This also allows each individual portion to be analysed. However, this method is not
scalable, as more portions means more features, leading to the curse of dimensionality.

\subsection{Experimental design}
The mouse movement, game statistic and itemization features were explored both
separately and together to identify their performance in player prediction. It was assumed
that in the dataset, all replays that were downloaded from the same account were
matches by the same player.
\begin{figure}
\centering
  \resizebox {1\columnwidth} {!} {
\pgfdeclarelayer{background}
\pgfdeclarelayer{foreground}
\pgfsetlayers{background,main,foreground}

\tikzstyle{data}=[draw, fill=blue!20, text width=4em, 
    text centered, minimum height=2em]
\tikzstyle{model} = [data, text width=4em, fill=red!20, 
  minimum height=2.5em, rounded corners]
\tikzstyle{puts}=[draw, fill=blue!20, text width=4em, 
    text centered, minimum height=2em]
\def\blockdist{3}
\newcommand{\newMoveModel}[3]{
    \node (#2-model) at (2,#3) [model] {#1 Model};
    \path (#2-model.west)+(-\blockdist,0) node (#2) [data] {#1 actions}; 
   
    \path (#2-model.east)+(\blockdist,0) node (#2-output) [data] {Predictions};
    
    \path [draw, ->] (#2.east) -- node {} (#2-model);
    \path [draw, ->] (#2-model.east) -- node{} (#2-output);
}

\newcommand{\newGameModel}[4]{
    \node (#2-model) at (2,#3) [model] {#1 Model};
    \path (#2-model.west)+(-\blockdist,0) node (#2) [data] {#4}; 
   
    \path [draw, ->] (#2.east) -- node {} (#2-model);
}
\def\blockdist{2}
\begin{tikzpicture}
\newGameModel{Attack}{ac}{4}{Attack actions}
\newGameModel{Move}{mc}{3}{Move actions}
\newGameModel{Spell}{sc}{2}{Spell actions}
\newGameModel{Stats}{stats}{1}{Game statistics}
\newGameModel{Items}{items}{0}{Itemization}

\path (sc-model.west)+(-4.5,0) node (game) [puts] {Match};
\path (sc-model.east)+(2,0) node (combine) [model] {MLP};
\path (combine.east)+(1.5,0) node (output) [puts] {Prediction};

\path [draw, ->] (game.20) -- node [above] {} (ac.180);
\path [draw, ->] (game.10) -- node [above] {} (mc.180);
\path [draw, ->] (game.0) -- node [above] {} (sc.180);
\path [draw, ->] (game.350) -- node [above] {} (stats.180);
\path [draw, ->] (game.340) -- node [above] {} (items.180);

\path [draw, ->] (ac-model.east) -- node [above] {} (combine.160);
\path [draw, ->] (mc-model.east) -- node [above] {} (combine.170);
\path [draw, ->] (sc-model.east) -- node [above] {} (combine.180);
\path [draw, ->] (stats-model.east) -- node [above] {} (combine.190);
\path [draw, ->] (items-model.east) -- node [above] {} (combine.200);

\path [draw, ->] (combine.east) -- node [above] {} (output);

\begin{pgfonlayer}{background}
  \path (ac.west |- ac.north)+(-0.5,0.5) node (a) {};
  \path (ac-model.south -| combine.east)+(+0.5,-0.5) node (b) {};
  \path (combine.east |- items-model.east)+(+0.5,-1) node (c) {};        
  \path[fill=yellow!20,rounded corners, draw=black!50, dashed] (a) rectangle (c);           
  \path (ac.north west)+(-0.2,0.2) node (a) {};        
\end{pgfonlayer}
\end{tikzpicture}
}
\caption{Classifying a single match with multiple different features.}

\label{fig:model-combination}
\end{figure}
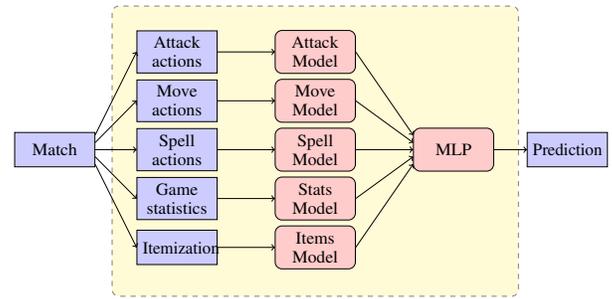

As the different features had different dimensions, each were evaluated with separate
models, with the results of each model combined using a simple multi-layer perceptron
with 1 hidden layer. Figure \ref{fig:model-combination} shows how matches split up
into each feature and model separately before being combined. A multi-layer perceptron
is used as the combination step so it can learn the relative weights for each feature,
rather than use them all equally in a voting scheme or set arbitrary weights to them.
Further, this approach is modular and allows features to be added or removed easily
to test the performance of different combinations of features.

Accuracy, precision and recall metrics were gathered using a 5-fold cross validation.
Accuracy measures the overall performance of a model by computing the
number of correct predictions over the total number of predictions.
Precision and recall and complementary metrics that measure the relevance
of predictions, with precision taking into account false positive results
and recall taking into account false negative results.
\begin{equation*}
\text{precision} = \frac{\text{true positives}}{\text{true positives} + \text{false positives}}
\end{equation*}

\begin{equation*}
\text{recall} = \frac{\text{true positives}}{\text{true positives} + \text{false negatives}}
\end{equation*}
Three different models were chosen based on models that previous work found
successful. The three models are logistic regression, random forest classifier
and multi-layer perceptron.

\section{Results}

\subsection{Player identification (experiment 1)}

\subsubsection{Mouse movement features}

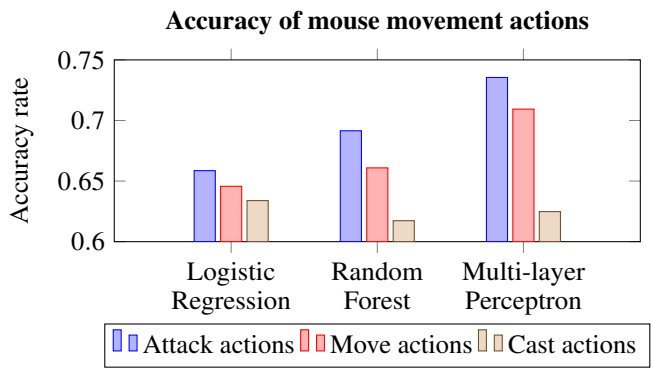
\begin{figure}
\centering
\begin{tikzpicture}
\newaxis{Accuracy of mouse movement actions}{Accuracy rate}{0.6}{0.75}{1\columnwidth}{4cm}{(0.5,-0.45)}{-1}
\addplot coordinates {(Logistic Regression,0.6586) (Random Forest,0.6915) (Multi-layer Perceptron,0.7355)};
\addplot coordinates {(Logistic Regression,0.6457) (Random Forest,0.6609) (Multi-layer Perceptron,0.7094)};
\addplot coordinates {(Logistic Regression,0.6339) (Random Forest,0.6173) (Multi-layer Perceptron,0.6248)};
\legend{Attack actions, Move actions, Cast actions}

\end{axis}
\end{tikzpicture}
\caption{Accuracy of individual types of complex mouse action. It can be seen that
  attack actions are the most useful for player prediction across all three models.}
\label{fig:mouse-individual}
\end{figure}

\begin{figure}
\centering
\begin{tikzpicture}
\newaxis{\textbf{Matches with mouse movements features}}{Metric rate}{0.1}{1.03}{0.9\columnwidth}{4cm}{(0.5,-0.45)}{-1}

\plotbargame{mouse}{accuracy}{Accuracy}
\plotbargame{mouse}{precision}{Precision}
\plotbargame{mouse}{recall}{Recall}

\end{axis}
\end{tikzpicture}
\caption{Results of classifying each match with all three types of mouse actions combined.}
\label{fig:mouse-combined}
\end{figure}
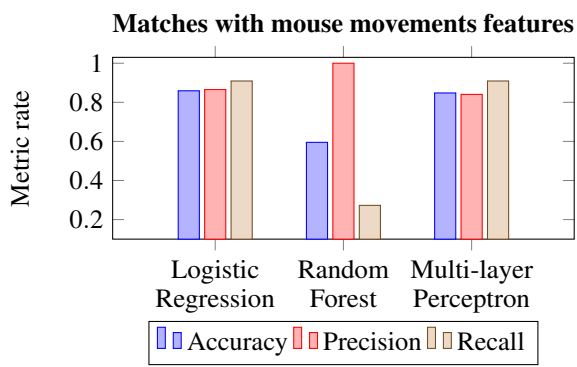
First, the mouse movement features were evaluated individually to give an
indication as to which type of complex mouse action is more informative for
player prediction. Figure \ref{fig:mouse-individual} shows that mouse movements
following an attack action were the most predictive, with movements following
a spell cast being the least predictive. It is important to note the
distribution of complex mouse actions are not equal, with an average of
20\%, 78\% and 2\% for attack, move and cast actions respectively. This directly
affects the number of data points available for training and testing.

Taking this into account, complex attack actions make a strong case for being the
most effective at predicting player behavior, as they are less common, but more
predictive than complex move actions. Complex move actions seem to be too common
to be a strong indicator for player behavior. They are sent in the order of a few
thousand commands in every match, due to how movement and positioning are fundamental
parts of Dota 2's gameplay. This commonality can lead to difficulty in distinguishing
between different players, especially when players are sending move commands multiple
times a second, making the mouse movement sequence very short. Complex spell actions
should follow the pattern for complex attack actions, since they are also rare and
unique compared to move actions. However, it seems their rarity (2\%) affects
their usefulness.

When looking at the mouse actions used together in figure \ref{fig:mouse-combined},
we can see the differences between each
machine learning model clearly. The random forest classifier has perfect precision
but low recall, indicating a lack of false positives predicted. From the test set,
we found 0 false positives and a large proportion of false negative predictions.
There are still true positive predictions, so the model is not predicting only negatives,
but is highly biased towards it. This suggests a specific player's (the positive
sample) behavior is difficult to differentiate from another player in most cases,
but when it is detected it is very obviously from the specific player.

The performance of logistic regression was quite noteworthy as it matches the neural
network in all three performance metrics. This is an improvement compared to using
individual complex mouse actions. This also shows that the combining network was able
to learn useful weights to apply to each of the three logistic regression models and
such a mixture of different complex mouse action types are more useful than each
mouse action alone.

\subsubsection{Game statistic features}
\begin{figure}
\centering
\begin{tikzpicture}
\newaxis{\textbf{Matches with game statistic features}}{Metric rate}{0.1}{1.03}{0.9\columnwidth}{4cm}{(0.5,-0.45)}{-1}

\plotbargame{stats}{accuracy}{Accuracy}
\plotbargame{stats}{precision}{Precision}
\plotbargame{stats}{recall}{Recall}

\end{axis}
\end{tikzpicture}
\caption{Results of classifying each match with game statistics.}
\label{fig:stats}
\end{figure}
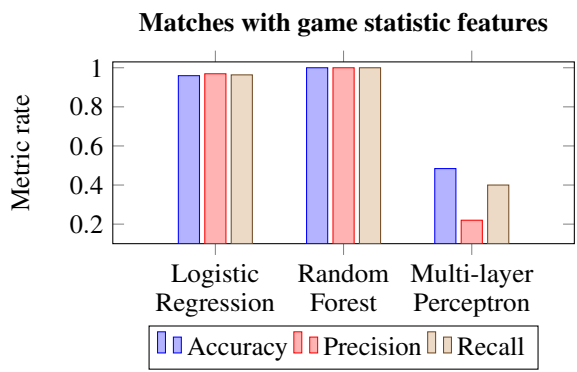

Using game statistic features gave a set of different results compared to using mouse
action features, with the neural network performing poorly and the other two models
achieving very high accuracies. As shown in figure \ref{fig:stats},
the random forest classifier had perfect results. To confirm this, we ran
further cross validation with varying values of $k$ up to 12 for additional trials.
It was found that the accuracy only fell twice, at $k = 2$ to 0.95 and at $k=8$ to 0.98.
This gives the strong performance merit, suggesting there exists a combination of rules
in a decision tree for player in-game statistics that easily identify players uniquely.

\subsubsection{Itemization features}
\begin{figure}
\centering
\begin{tikzpicture}
\newaxis{\textbf{Accuracy of itemization features}}{Accuracy rate}{0.8}{1}{\columnwidth}{4cm}{(0.5,-0.45)}{2}

\plotbargame{items-hashed}{accuracy}{Items hashed}
\plotbargame{items-onehot}{accuracy}{One-hot encoding}
\plotbargame{items-starting}{accuracy}{Starting items}
\plotbargame{items-select}{accuracy}{Boots only}
\end{axis}
\end{tikzpicture}
\caption{Results of classifying each match with itemization features.}
\label{fig:items}
\end{figure}
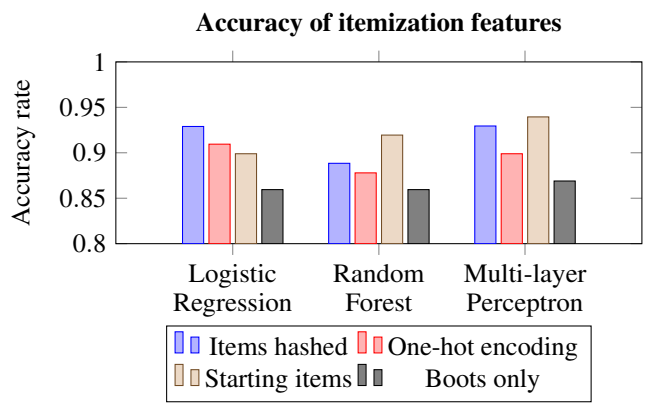

Each encoding method of itemization gave different performances, with the random forest
classifier and multi-layer perceptron giving the same pattern for each encoding method,
as shown in figure \ref{fig:items}.
Firstly, encoding with boots only data had the lowest accuracy of all the encoding methods.
This makes sense as it uses strictly less items compared to other encoding methods.
This represented a trade-off for computation of features and accuracy.
For comparison, the one-hot encoding of only
starting items gets 4\%, 6\% and 7\% more accuracy for each model respectively, but uses
186 features compared to the 42 features of encoding only boots data.

We also found a small improvement from using starting items only compared to using all items
with one-hot encoding. Recall that the encoding of all items is sampled twice in a match,
once at the beginning and once at the end.
The close performance of the two encoding methods suggests a few things. Firstly, the
massive increase in binary features does not heavily impact performance. The encoding of
all items contains about 9 times the number of features (281 items vs 31 items)
but the accuracy deviates by less than 5\%.
Secondly, though the starting items do not have the second sample of items at the end of
the match, it still predicts better in two of the three machine learning models.
This shows the items locations of at the start of a match is more useful in player prediction
than at the end of a match. This makes sense from the game's perspective, as items a player
buys throughout a match is often dependent on the strategy, composition and context of their
team, the opposing team and how well they are doing, so there is more variance in end of match
items compared to the start of matches.
Further, many matches end before a player finishes buying all items they need, so item
locations recorded at the end of a match could be temporary and different from other matches
by the same player, depending on the context of the match. 

\subsubsection{Combined features}
Combining the features together gave mixed results in most cases. There were some improvements
in specific cases, but there were also cases where performance decreased.

Logistic regression was able to predict well in general, with the combination of
game statistics and boots achieving 99\% accuracy.
There was also an interesting anomaly in combining mouse movement, game
statistics and one-hot encoding of all items that had a drastic reduction in accuracy. This
may be due to the combination of one-hot encoding of items and mouse movement being a bad
combination, as it was the only other combination that had lower than 90\% accuracy.

\begin{table}
\centering
\begin{tabular}{ p{0.4\columnwidth} | >{\centering\arraybackslash}p{0.2\columnwidth} >{\centering\arraybackslash}p{0.12\columnwidth} >{\centering\arraybackslash}p{0.2\columnwidth}}
& \multicolumn{3}{c}{\textbf{Models}} \\ & & & \\
\textbf{Features} & Logistic Regression & Random Forest & Multi-layer Perceptron \\
Mouse + Stats & 0.959 & 0.597 & \textcolor{red}{0.738} \\
Mouse + Items hashed & 0.919 & 0.646 & 0.909 \\
Mouse + Items one-hot & 0.888 & 0.606 & 0.939 \\
Mouse + Starting items & 0.918 & 0.617 & 0.919 \\
Mouse + Boots only & 0.911 & 0.857 & 0.908 \\
Stats + Items hashed & 0.969 & \textcolor{red}{0.990} & 0.868 \\
Stats + Items one-hot & 0.969 & \textcolor{red}{0.979} & 0.939 \\
Stats + Starting items & 0.959 & \textcolor{red}{1.0} & 0.929 \\
Stats + Boots only & \textcolor{red}{0.99} & \textcolor{red}{0.959} & 0.859 \\
Mouse + Stats + Items hashed & 0.949 & 0.636 & \textcolor{red}{0.808} \\
Mouse + Stats + Items one-hot & \textcolor{red}{0.726} & 0.769 & \textcolor{red}{0.919} \\
Mouse + Stats + Starting items & 0.959 & 0.586 & \textcolor{red}{0.929} \\
Mouse + Stats + Boots only & 0.959 & 0.615 & \textcolor{red}{0.928} \\
\end{tabular}
\caption{Accuracy of each model on each feature combination. The accuracies highlighted in red are of particular interest for discussion.}
\label{tbl:game-combination}
\end{table}

The random forest classifier showed reduced accuracy in most combinations using mouse movements
compared to using the features alone. For example, the various itemization encoding methods
had about 85-90\% accuracy and game statistics had close to 100\%. However, when any of these
features are combined with the mouse movement data, the accuracy drops. It can also be seen
in table \ref{tbl:game-combination} that the combination of itemization and game statistics
do well, so it is only the mouse movement features that is an issue
for this model. Recall that the precision for using mouse movement features in the
random forest classifier was extremely high with low recall, which reduced its accuracy.
The lack of accuracy even when combined with accurate features such as game statistics
shows that it is not enough to combine all features together, and bad features add a lot of
noise to the random forest model.

Finally, the multi-layer perceptron was able to combine features together better.
Though it performed poorly using only statistics, the addition of mouse movements and itemization
features were able to help boost its accuracy. Interestingly, the better individually performing
item encoding methods were not always better than others when combined with other features.
For example, using mouse features with the hashed items and mouse features with boots only
gave similar accuracy, even though the hashed encoding was more accurate alone compared to
the boots only feature.

\subsection{Same player identification (experiment 2)}

\subsubsection{Mouse movement features}
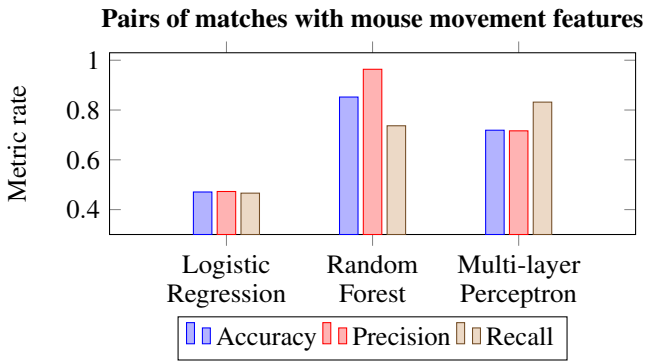
\begin{figure}
\centering
\begin{tikzpicture}
\baraxis{\textbf{Pairs of matches with mouse movement features}}{Metric rate}{0.3}{1.03}{1\columnwidth}{4cm}{(0.5,-0.45)}{1.7cm}{-1}

\plotbarpair{1}{1}{mouse}{accuracy}{Accuracy}
\plotbarpair{1}{1}{mouse}{precision}{Precision}
\plotbarpair{1}{1}{mouse}{recall}{Recall}

\end{axis}
\end{tikzpicture}
\caption{Accuracy, precision and recall of using mouse movement features only when classifying pairs of matches.}
\label{fig:mouse-pair}
\end{figure}

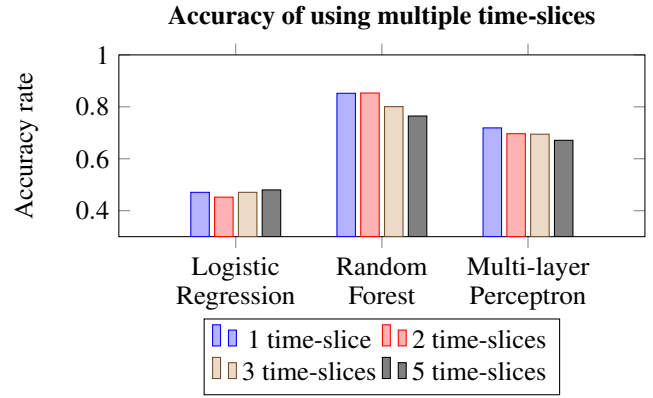
\begin{figure}
\centering
\begin{tikzpicture}
\baraxis{\textbf{Accuracy of using multiple time-slices}}{Accuracy rate}{0.3}{1}{1\columnwidth}{4cm}{(0.5,-0.45)}{1.7cm}{2}

\plotbarpair{1}{1}{mouse}{accuracy}{1 time-slice}
\plotbarpair{2}{2}{mouse}{accuracy}{2 time-slices}
\plotbarpair{3}{3}{mouse}{accuracy}{3 time-slices}
\plotbarpair{5}{5}{mouse}{accuracy}{5 time-slices}

\end{axis}
\end{tikzpicture}
\caption{Accuracy of mouse movement features when a match is split into a different number of time-slices. More time-slices means a greater number of features, as the featureset is repeated for each time-slice.}
\label{fig:mouse-pair-slices}
\end{figure}

Figure \ref{fig:mouse-pair} shows the performance of the three machine learning methods.
In general, logistic regression performs much worse for this problem,
regardless of how the match is split into time-slices.
This shows the non-linear nature of this problem compared to the first one.
Splitting the match into multiple time-slices doesn't help or hinder the accuracy for
logistic regression, which shows it is unlikely an issue of too many features affecting the
performance, otherwise a greater fall in accuracy would be expected from
figure \ref{fig:mouse-pair-slices}.

The overall accuracies of all three models were also lower compared to the first problem,
which shows the difficulty of this problem in comparison, when there is not a
small and limited pool of players.

\begin{figure}
\centering
\begin{tikzpicture}
\begin{axis}[
    title={\textbf{Accuracy of individual time-slices}},
    grid=major,
    xmax=5,xmin=1,
    width=1\columnwidth,
    height=4cm,
    xlabel={Time-slice of match},
    ylabel={Accuracy rate},
    legend style={at={(0.5,-0.5)},anchor=north,legend columns=2}
]

\plotline{Logistic Regression}{mouse}{accuracy}{Logistic Regression}
\plotline{Random Forest}{mouse}{accuracy}{Random Forest}
\plotline{Multi-layer Perceptron}{mouse}{accuracy}{Multi-layer Perceptron}

\end{axis}
\end{tikzpicture}
\caption{Accuracy for mouse movement features on individual portions of a match.}
\label{fig:mouse-pair-individual-slices}
\end{figure}
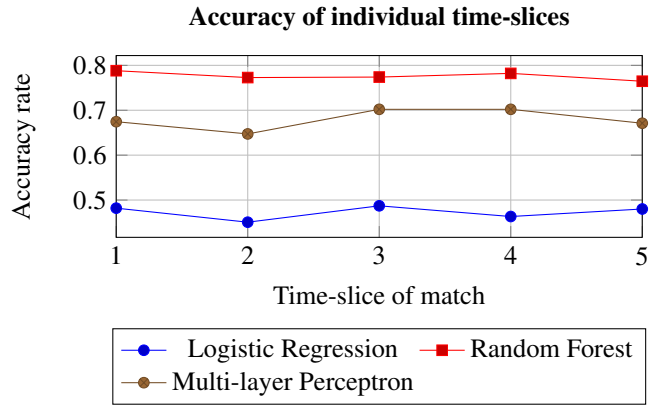

Moreover, using additional time-slices caused less accurate predictions, which suggests the
existence of noise when the featureset is repeated for each time-slice. Both the random
forest classifier and multi-layer perceptron lost accuracy as a match was sliced into
more parts and used together. This shows that although slicing gives more fine-grained
data, the additional noise is detrimental to player prediction. We also found
that each individual time-slice has little effect on the accuracy of predictions,
as figure \ref{fig:mouse-pair-individual-slices} show little performance difference between
each individual time slice.
This is rather unexpected, since the strategy and gameplay in different phases
of a Dota 2 game change, so one would expect the mouse movement behavior of players
to change as well.

\subsubsection{Game statistic features}

Figure \ref{fig:stats-correlation} shows the correlation matrix between pairs of matches
by the same player. We can see how each statistic feature of one match is correlated
to the statistic feature of another match from the same player. For instance, the
CS, denies, gold per minute, XP per minute and CS per minute statistics are well
correlated with each other, even across different matches. It would be expected for
these numbers to be correlated in the same match, as creeps directly affect the gold and
XP gained by heroes. However, seeing the statistics continue this trend across different
matches suggest they are also tied to specific player behavior.

It should be noted that the correlation matrix is mirrored across the diagonal. This is
a result of the combination step, where matches are combined into pairs of
\textit{\{match0, match1\}} and \textit{\{match1, match0\}} and should not be taken
as a meaningful result. Due to the way the \texttt{clarity} parser runs through matches
chronologically, a bug in the library prevented us from splitting up a match into
time-slices like the mouse movement feature for game statistics.

Once again, the random forest classifier performs the best when using the game statistic
features, showing a strong correlation between game statistics and player prediction
using decision tree based models. This seems to be due to decisions tree being able
to determine ranges for statistics for player prediction. For example if one statistic
is always in a similar range across multiple games, it is likely to be the same
player.

\subsubsection{Itemization features}
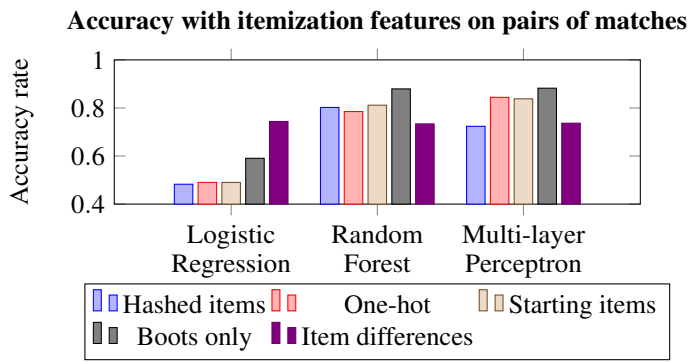
\begin{figure}
\centering
\begin{tikzpicture}
\baraxis{\textbf{Accuracy with itemization features on pairs of matches}}{Accuracy rate}{0.4}{1}{1\columnwidth}{3.5cm}{(0.5,-0.55)}{2cm}{3}

\plotbarpair{1}{1}{items-hashed}{accuracy}{Hashed items}
\plotbarpair{1}{1}{items-onehot}{accuracy}{One-hot}
\plotbarpair{1}{1}{items-starting}{accuracy}{Starting items}
\plotbarpair{1}{1}{items-select}{accuracy}{Boots only}
\plotbarpair{1}{1}{items-diff}{accuracy}{Item differences}

\end{axis}
\end{tikzpicture}
\caption{Accuracy for game itemization features when classifying if two matches belong to the same player.}
\label{fig:items-pair}
\end{figure}
Figure \ref{fig:items-pair} shows the accuracy with different itemization encodings and features.
For these features, using only boots was the best indicator for random forest and
multi-layer perceptron, while the item differences were best for the logistic regression model.
Once again, logistic regression did not have good results compared to the other models.
The only exception is the item difference method of encoding. This is a bit of an anomaly,
as the trend suggests logistic regression is not suitable for this classification task. 

The difference between the item difference encoding and all other features is that features
are not duplicated for the pair of matches. For all other features, the features for both
matches are used together, to see if features across the two matches may be correlated.
The game statistic features shown earlier display this kind of correlation.
However for the item difference, the presence or absence of the same item on an inventory
slot is encoded and so there is no duplication of features. This contrast reduces the
complexity of the feature, which allowed the logistic regression model to perform
comparably to the other two models where all other features could not.

The one-hot encoding of all items performed similarly to the one-hot encoding of only 
starting items. This was surprising because the one-hot encoding of all items contained
250 more items, which translates to 1500 more binary features. As was the case in the last
experiment, the increased number of binary features had little effect on the performance
of the models. The gap between the two methods is smaller than previously, which means
just using starting items to differentiate between two players is more difficult than
identifying a single player. This suggests a more diverse pool of players leads
to more similarity in starting items between any two players. 

\subsubsection{Combined features}

\begin{figure}
  \centering

\begin{tikzpicture}
\begin{axis}[
    ybar,
    title={\textbf{Combining features a random forest model}},
    width=1\columnwidth,
    height=4cm,
    bar width=0.6em,
    legend style={at={(0.5,-0.7)},anchor=north,legend columns=-1},
    enlarge x limits=0.2,
    x tick label style={align=center,text width=1.1cm},
    symbolic x coords={Starting items only, Mouse + items, Stats + items, Mouse + stats, Mouse + stats + items},
    xtick=data,
    ylabel={Metric rate},
]
\addplot table [x=features, y=accuracy, col sep=comma] {
numSplits,split,accuracy,precision,recall,model,features
1,1,0.811532212460573,0.8271052919831957,0.7941936184067723,Random Forest,Starting items only
1,1,0.9135779793479282,0.9102883801009354,0.9225173648795313,Random Forest,Mouse + items
1,1,0.9102086605455826,0.9537679752573119,0.8635228999348816,Random Forest,Stats + items
1,1,0.950077781845275,0.9502363396489525,0.9520675059691772,Random Forest,Mouse + stats + items
1,1,0.9322103786068515,0.9219059921138925,0.9469068808335143,Random Forest,Mouse + stats

};
\addlegendentry{Accuracy}

\addplot table [x=features, y=precision, col sep=comma] {
numSplits,split,accuracy,precision,recall,model,features
1,1,0.811532212460573,0.8271052919831957,0.7941936184067723,Random Forest,Starting items only
1,1,0.9135779793479282,0.9102883801009354,0.9225173648795313,Random Forest,Mouse + items
1,1,0.9102086605455826,0.9537679752573119,0.8635228999348816,Random Forest,Stats + items
1,1,0.950077781845275,0.9502363396489525,0.9520675059691772,Random Forest,Mouse + stats + items
1,1,0.9322103786068515,0.9219059921138925,0.9469068808335143,Random Forest,Mouse + stats
};
\addlegendentry{Precision}

\addplot table [x=features, y=recall, col sep=comma] {
numSplits,split,accuracy,precision,recall,model,features
1,1,0.811532212460573,0.8271052919831957,0.7941936184067723,Random Forest,Starting items only
1,1,0.9135779793479282,0.9102883801009354,0.9225173648795313,Random Forest,Mouse + items
1,1,0.9102086605455826,0.9537679752573119,0.8635228999348816,Random Forest,Stats + items
1,1,0.950077781845275,0.9502363396489525,0.9520675059691772,Random Forest,Mouse + stats + items
1,1,0.9322103786068515,0.9219059921138925,0.9469068808335143,Random Forest,Mouse + stats
};
\addlegendentry{Recall}

\end{axis}
\end{tikzpicture}
\caption{Performance increases as different features are combined with the starting item feature for the random forest classifier.}
\label{fig:pair-combined-rf}
\end{figure}
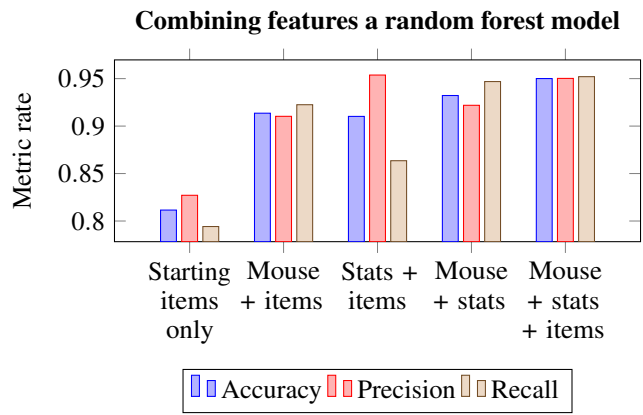

\begin{figure*}
\centering
  \includegraphics[width=1.35\columnwidth]{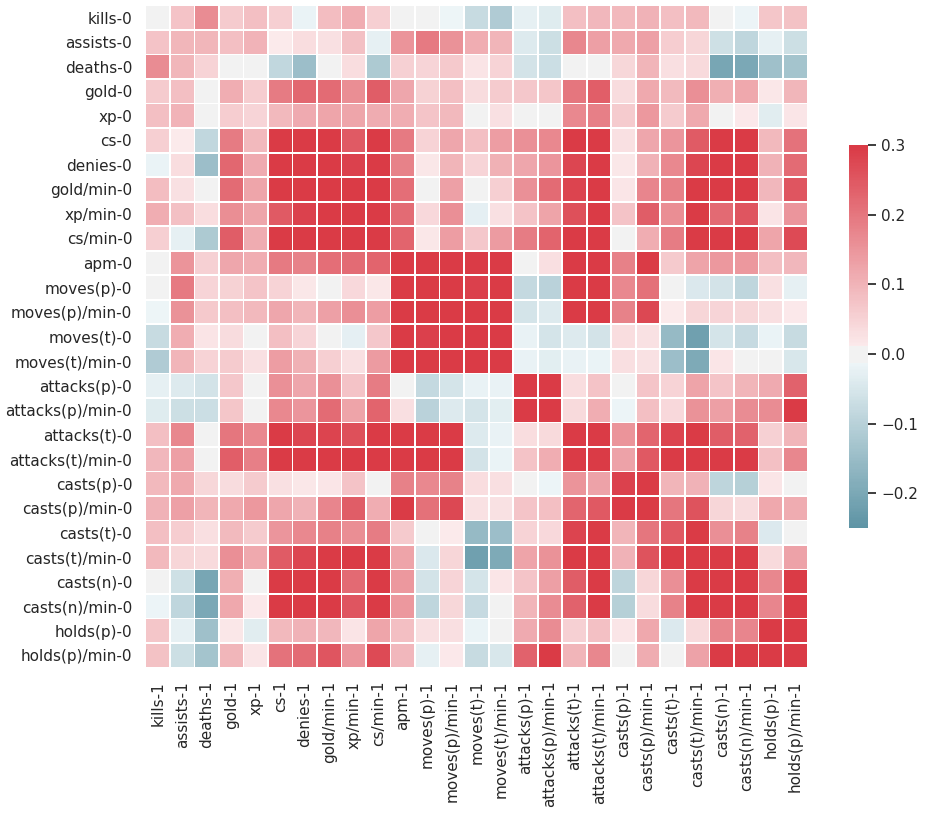}
  \caption{Correlation matrix for pairs by the same players}
\label{fig:stats-correlation}
\end{figure*}

Strong results were found for combining features together, unlike the first experiment.
All three models were able to find improvements from using the different features
together. For instance, the logistic regression model could not accurately predict
if two matches belonged to the same player for all features except for the item
difference feature. However, when combining the item difference feature with
the poor performing mouse movement and game statistic features,
we were able to get up to 10\% higher accuracy on certain combinations.
For the multi-layer perceptron, we found precision generally increased with more
featuresets at the cost of recall. This led to a smaller increase in accuracy of
5-7\%. The high precision is more beneficial compared to high recall because
less false positives means predictions for two matches belonging to the same
player are more unlikely to be wrong, even if we may miss some predictions
due to false negatives.

We found the random forest classifier was able to achieve the highest accuracies
in this problem when multiple features are combined together.
Figure \ref{fig:pair-combined-rf} shows the improvement we found. The model was able to go
from 81\% correct predictions using only starting items to 91\%-95\% correct predictions
from a combinations of starting items and other features. The highest 95\% accuracy came
from a combination of starting items, game statistics and mouse movement features.

\section{Future work}

\subsubsection{Fine grained information on mouse movement features}
Although we have been successful in using mouse movement features to predict players with good
results, there are still many details that are unknown. For example, although we take
a number of attributes for each mouse movement sequence recorded, the context such as
target entity or current hero properties are not recorded. Including the context for each
mouse movement would allow us to learn if certain actions under certain circumstances
are more indicative of player behavior.

\subsubsection{Multiple heroes from the same player}
In this paper, we focused our research on the exploration and combinations of features
and models for player prediction so we kept the hero used constant throughout the
experiments. The same methodology is likely to work on different heroes, but it
is not known how it would work in a dataset that contained multiple heroes. This
is because the difference in play-style of different heroes would interfere with the
difference in player behavior, so it would be interesting to see how this work could
be extended to a multi-class classification of this case.

\subsubsection{Deep Learning}
Our work primarily used traditional machine learning models such as logistic regression
and random forest classifiers. With the increasing popularity and performance of deep
neural networks, it would be interesting to compare how a accurate a deep learning
model could predict players compared to these traditional machine learning models.

\subsubsection{Automated detection tool}
Finally, because we have found strong results for classifying with random forest
classifiers, our models can be applied to create a real world tool for automatic
detection. The tool can be prototyped to accept replays and return predictions.
Successful prototyping with a high accuracy and high precision can lead to the
tool being used in online tournaments as a method to prevent cheating.
The creation of such a tool will also help to refine the model by exposing
it to a much larger variety of players and matches.

\section{Conclusion}
This work explores the extraction and use of various features from Dota 2 match
replays, in combination with mouse tracking techniques, to predict who the player is based on their in-game behavior.
We explore two approaches to this problem. Firstly, we predict identity from
specific player from a pool of players. For this approach, we find that the use of game
statistics with a random forest classifier can achieve high prediction rates
but it could not combine poor features with good features without losing a lot of
accuracy.

Our second, more universal approach centres around matching (or otherwise) players from two different matches from the set of all players. This approach
is more generalizable to the large player-base. We find the combination of game starting items,
mouse movements and game statistics using a random forest classifier can produce an accuracy of 95\%.

\bibliographystyle{SIGCHI-Reference-Format}
\bibliography{sample}


\end{document}